\documentclass{article}

\usepackage{microtype}
\usepackage{graphicx}
\usepackage{subcaption}
\usepackage{booktabs}
\usepackage{multirow}
\usepackage{array}
\usepackage{hyperref}
\usepackage[accepted]{icml2026}
\usepackage{amsmath}
\usepackage{amssymb}
\usepackage{xcolor}
\usepackage{colortbl}
\usepackage{enumitem}

\newcommand{\mabv}{MAB-v3}

\icmltitlerunning{World Feedback for Clinical Agents}

\begin{document}

\twocolumn[
\icmltitle{World Feedback for Clinical Agents:\\
Diagnosing RL in FHIR Environments}

% \begin{icmlauthorlist}
% \icmlauthor{Anonymous}{anon}
% \end{icmlauthorlist}
% \icmlaffiliation{anon}{Anonymous Institution}
% \icmlcorrespondingauthor{Anonymous}{anonymous@anon.org}

\begin{icmlauthorlist}
\icmlauthor{Ananya Mantravadi}{centific}
\icmlauthor{Harshit Rajgarhia}{centific}
\icmlauthor{Prasanna Desikan}{centific}
\icmlauthor{Abhishek Mukherji}{centific}
\end{icmlauthorlist}

\icmlaffiliation{centific}{Centific Global Solutions, Inc.}

\icmlcorrespondingauthor{Ananya Mantravadi}{ananya.mantravadi@centific.com}
\icmlkeywords{reinforcement learning, clinical AI, FHIR, world feedback, GRPO}
\vskip 0.3in
]

\printAffiliationsAndNotice{}

% ─────────────────────────────────────────────────────────────────────────────
\begin{abstract}
Clinical protocol-execution tasks---checking a lab value, applying a threshold,
placing a correctly structured FHIR order---are natural candidates for RL from
world feedback: once clinical SMEs encode decision logic into a verifier, that
verifier grades unlimited rollouts without per-episode annotation. But applying
RL requires a sound feedback channel and sufficient base capability. We audit
MedAgentBench v1/v2, find a 41.7\% silent-finish ceiling that makes inaction
the RL dominant strategy, and construct \textbf{MedAgentBench-v3 (MAB-v3)}
(508 tasks, 8.9\% ceiling). Training Qwen3-8B exposes two structural barriers:
a \emph{capability ceiling} (10/20 task types have 0\% base performance, zero
gradient) and a \emph{format-knowledge barrier} (3/20 types require exact
clinical codes undiscoverable by exploration). Pure RL reaches 18.2\% pass@1
vs.\ 34.1\% for rule-based SFT; the 15.9~pp gap is attributable entirely to
these barriers. A decision/format-knowledge/lookup taxonomy predicts RL
learnability and prescribes the fix: SFT to inject codes, RL to learn
conditionals.
\end{abstract}

% ─────────────────────────────────────────────────────────────────────────────
\section{Introduction}
\label{sec:intro}

A large class of clinical tasks involves \emph{protocol execution}: given a
known decision rule, the agent retrieves a lab value, applies the threshold,
and if triggered places a correctly structured FHIR order. These are
administrative workflow tasks---not replacement of physician judgment, but
execution of standing orders~\citep{jiang2024mab,lee2025fhiragentbench,bedi2026healthadminbench}.
MedAgentBench v1 and v2, defining the 20 task types studied here, were
constructed with clinical teams who validated each workflow against real EHR
practice~\citep{jiang2024mab,mab2025v2}.

\paragraph{Why RL from world feedback:}
Protocol correctness is verifiable: once a clinical SME encodes the decision
logic into a verifier, that verifier grades every subsequent rollout
automatically. This is qualitatively different from
RLHF~\citep{christiano2017deep,ouyang2022rlhf}: SME effort is front-loaded
into environment design rather than spent labeling individual episodes. The
alternative---supervised SFT demos---requires manually coding every clinical
rule, is biased toward action-branch instances, and must be regenerated when
protocols change. RL from world feedback avoids this: the agent explores the
environment and receives feedback from the verifier, with only a single update
needed when a protocol changes.

\paragraph{What our experiments reveal:}
Applying RL here is non-trivial. Two structural barriers limit a naive
approach. First, the feedback channel must be clean: MedAgentBench v1/v2 had
a 41.7\% silent-finish ceiling (41.7\% of tasks pass with no tool use),
making inaction the RL dominant strategy. GRPO~\citep{shao2024grpo} on
uncorrected MAB-v2 converged to 0\% action-branch pass. We construct
\textbf{MedAgentBench-v3 (\mabv{})} (508 tasks, 8.9\% ceiling) to fix this.
Second, even on a clean benchmark, 3 of 20 task types require exact clinical
codes (SNOMED, NDC) undiscoverable by exploration---flat reward landscape---and
10 types have zero base capability, yielding zero gradient. These two barriers
explain the 15.9~pp gap: SFT (34.1\%) injects codes and format; pure RL
(18.2\%) cannot. The approaches are complementary, and SFT+RL is the
prescription our results motivate.

\paragraph{Contributions:}
\begin{itemize}[leftmargin=1.2em,topsep=2pt,itemsep=1pt]
\item \textbf{\mabv{} + environment} (Section~\ref{sec:benchmark},
  \ref{sec:envdesign}): Corrected 508-task benchmark (silent-finish ceiling
  41.7\%$\to$8.9\%) with a self-contained world feedback environment: offline
  FHIR server, auditable rule-based verifier, and deliberate reward shaping
  for conditional clinical behavior.
\item \textbf{Structured diagnosis of RL limitations}
  (Section~\ref{sec:methods}--\ref{sec:results}): Base 16.6\%, SFT 34.1\%,
  pure RL 18.2\%. The 15.9~pp SFT/RL gap is attributable to format-knowledge
  and capability-ceiling failures, not to the RL algorithm or reward design.
\item \textbf{Task taxonomy} (Section~\ref{sec:analysis}): Decision /
  format-knowledge / lookup framework, validated by per-type results, that
  predicts RL learnability from first principles and prescribes SFT+RL as the
  right combination for mixed-structure clinical benchmarks.
\end{itemize}

% ─────────────────────────────────────────────────────────────────────────────
\section{Related Work}
\label{sec:related}

\paragraph{FHIR agent evaluation:}
\citet{jiang2024mab} introduced MedAgentBench, grounding clinical agent
evaluation in FHIR tool use; MAB v2~\citep{mab2025v2} extended it with 300
tasks. Neither examines benchmark validity as a training signal.
FHIR-AgentBench~\citep{lee2025fhiragentbench} targets factual retrieval from
MIMIC-IV rather than clinical action.
HealthAdminBench~\citep{bedi2026healthadminbench} documents a subtask/task
reliability gap parallel to the action/aggregate divergence we quantify.

\paragraph{RL from verifiable non-human feedback:}
Outcome-based RL with deterministic verifiers has driven large gains in
mathematics~\citep{shao2024grpo,deepseek2025r1}. Our contribution is applying
this paradigm to clinical environments and showing that task structure
determines whether the verifier provides learnable signal---a distinction
absent from homogeneous domains like arithmetic. Multi-task gradient
dominance~\citep{grpomultitask2024,mtgrpo2025} is a known failure mode we
address via per-task advantage normalization.

% ─────────────────────────────────────────────────────────────────────────────
\section{MedAgentBench-v3: Restoring the World Feedback Signal}
\label{sec:benchmark}

\subsection{Why the Original Benchmark Breaks RL}

Two properties combine to make inaction the dominant RL strategy on MAB-v2.

\textbf{Branch imbalance:}
Four v2 task types have 70--97\% of instances on the no-action branch due to
cohort composition. RL rewards the most common outcome; on these types,
abstaining is correct the vast majority of the time.

\begin{figure}[t]
  \centering
  \includegraphics[width=\columnwidth]{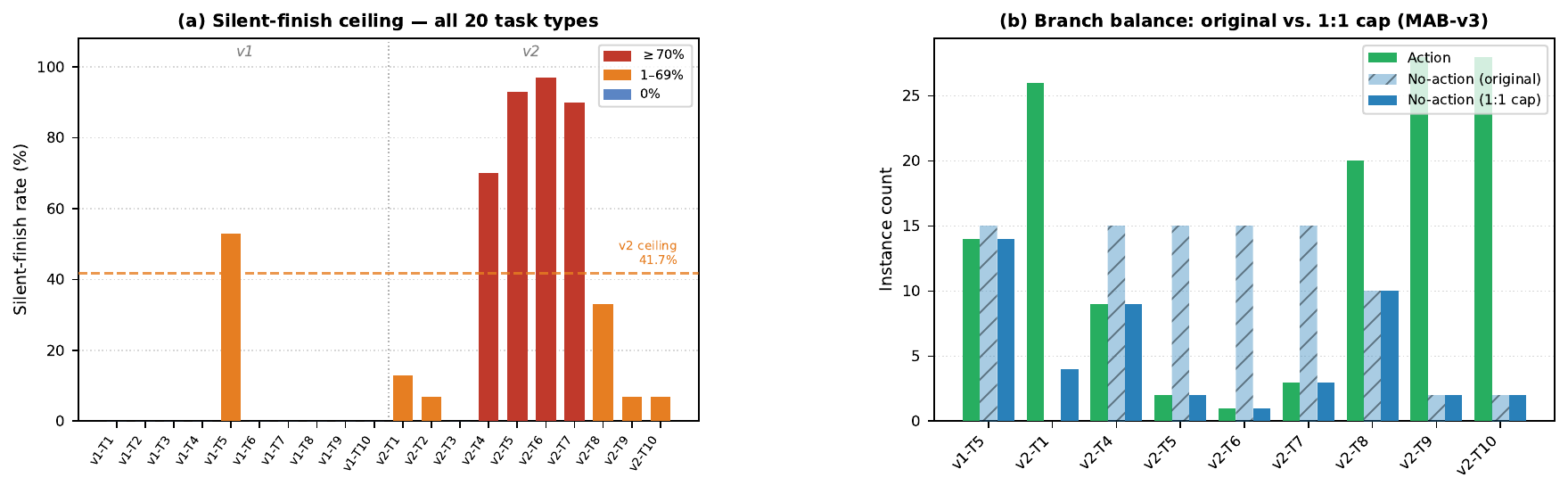}
  \caption{Branch imbalance before (hatched) and after (solid) the 1:1 cap in
    \mabv{}, and the resulting fall in the silent-finish ceiling from 41.7\% to
    8.9\%. The original imbalance made inaction the RL dominant strategy.}
  \label{fig:main}
\end{figure}

\textbf{Silent-finish ceiling:}
Running a null agent (immediate \texttt{finish([])}, no tool calls) on all 600
MAB-v2 tasks yields 41.7\% pass. This is not a bug in any single task---many
patients genuinely need no intervention---but it creates a dominant RL
strategy: ignoring FHIR data and abstaining scores 41.7\% before any clinical
action is learned. On uncorrected MAB-v2, GRPO converged to 0\% action-branch
pass within 200 steps (Figure~\ref{fig:main}).

\subsection{Additional Evaluation Failures}

\textbf{Undocumented format requirements:}
Graders for v1-T5, v1-T9, and v2-T3 enforce format conventions absent from
the task context: bare-string route fields, tiered dose formulas, and
two-element return arrays. Clinically correct but incorrectly formatted
responses fail systematically. We add the missing documentation to task
context.

\textbf{Wall-clock bug:}
The v2-T1 grader calls \texttt{datetime.now()} as the CT follow-up reference
date, misclassifying 4 of the 30 v2-T1 task instances as action-required when run in
2025--2026. We freeze the timestamp to \texttt{2023-11-13T10:15:00+00:00}.

\subsection{\mabv{} Construction}

Four corrections: (1)~context patches (v1-T5, v1-T9, v2-T3); (2)~fixed
timestamp (v2-T1); (3)~silent-finish labeling for all 600 tasks;
(4)~1:1 branch balance cap per (corpus, task type).

\textbf{Result.} \mabv{} draws from two source corpora---MAB v1 (original
300 tasks) and MAB v2-new (300 new tasks)---spanning 20 task types across $\approx$100 real anonymized patients
(30 task instances per type, drawn from the same patient pool). After curation it contains \textbf{508 tasks}: 463
action-required, 45 no-action. Silent-finish ceiling: \textbf{8.9\%}, down
from 41.7\%. The v1/v2-new split is preserved throughout evaluation to
distinguish the two distinct sets of clinical workflows.

% ─────────────────────────────────────────────────────────────────────────────
\section{Methods}
\label{sec:methods}

\subsection{Environment Design}
\label{sec:envdesign}

\begin{figure}[t]
  \centering
  \includegraphics[width=\columnwidth]{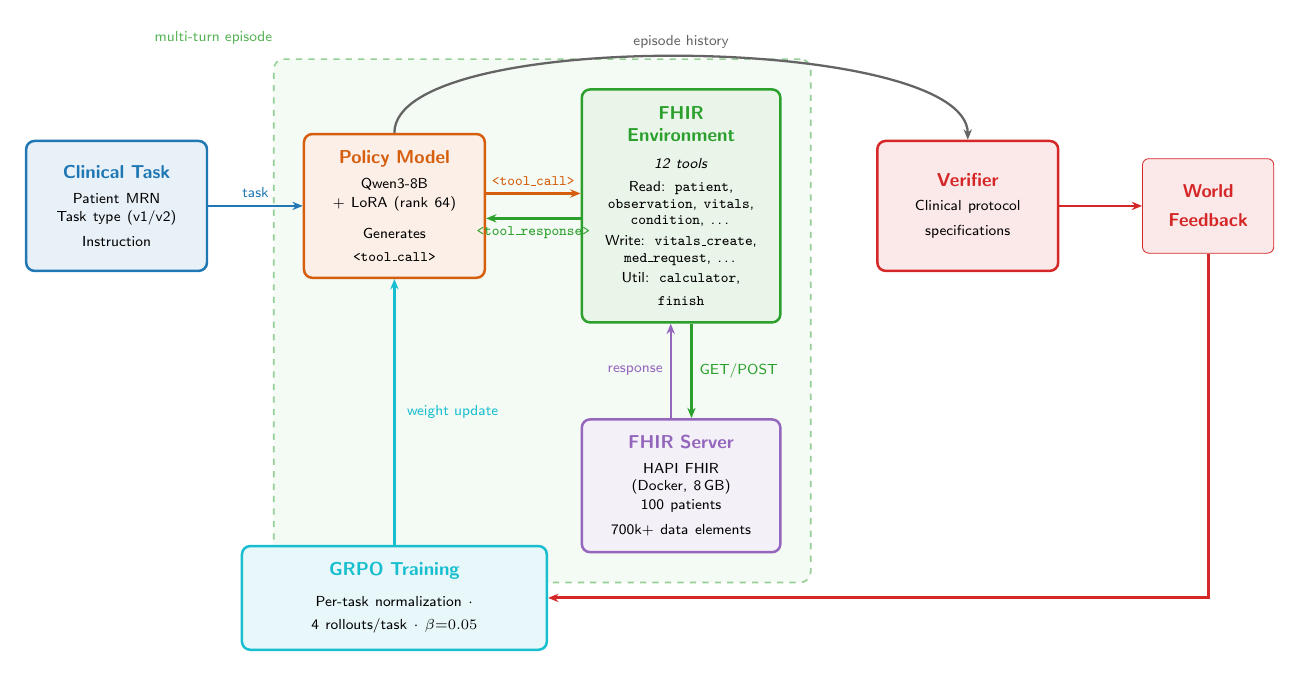}
  \caption{World feedback loop for clinical FHIR agent training. The policy
    model (Qwen3-8B + LoRA) generates \texttt{<tool\_call>} actions; the FHIR
    environment executes them against an offline server snapshot and returns
    deterministic responses. At episode end, a rule-based verifier grades the
    agent's actions against clinical protocol specifications and returns reward
    $r$---no per-episode human judgment involved (SME effort is concentrated
    in the one-time verifier design). GRPO uses this world feedback signal to
    update model weights. The multi-turn episode box (dashed) shows the
    interact-observe loop; the training loop closes through GRPO at bottom.}
  \label{fig:arch}
\end{figure}

Figure~\ref{fig:arch} shows the complete world feedback loop. A well-designed
RL environment should ensure that score variations reflect the capability being
measured, not environment instability, verifier bugs, or reward shortcuts. Our
benchmark audit (Section~\ref{sec:benchmark}) reveals that MAB-v2 fails this
on multiple dimensions. We document three design principles \mabv{} addresses
and two failure modes we observed in practice.

\textbf{Reproducible, deterministic episodes:}
The FHIR environment runs against a fixed snapshot of an HAPI FHIR server
covering $\approx$100 real anonymized patients. Every rollout is fully deterministic: the same query always
returns the same response, there is no server state, and episodes complete in
under one second. This is a prerequisite for the thousands of rollouts RL
requires and eliminates environment instability as a confound---a known failure
mode where environment instability causes failures unattributable to model
capability.

\textbf{Reward shaping for conditional behavior:}
The world feedback signal decomposes as:
\begin{equation}
r = r_{\text{terminal}} + r_{\text{action}} + r_{\text{penalty}},
\end{equation}
where $r_{\text{terminal}} = 1.0$ if the grader passes; $r_{\text{action}}
\in \{0.10, 0.25\}$ gives partial credit for correct resource type and POST
structure; $r_{\text{spurious}} = -0.15$ penalizes an off-target POST on a
no-action task; $r_{\text{skip}} = -0.20$ penalizes finish with no tool use.
The partial credit and $r_{\text{spurious}}$ are \emph{deliberate design
choices}, not natural properties of the clinical world. Without partial credit,
the reward landscape is flat until the exact correct order is placed, making
early exploration unrewarding. Without $r_{\text{spurious}}$, the model learns
to always act since action always earns partial credit. These two components
together create a gradient for both the action and the conditional structure.

\textbf{Auditable grader:}
The rule-based verifier (1,340 lines implementing medical protocol
specifications across all 20 task types) is itself the feedback provider: it
queries the same FHIR environment the agent uses, then checks POST payloads
against clinical criteria. Every failure is traceable to a specific criterion.
This auditability is what lets us identify the wall-clock bug and undocumented
format requirements as grader failures rather than model capability gaps.

\textbf{Format consistency:}
Training, evaluation, and SFT demonstration generation all use the
\texttt{<tool\_call>} named-function interface---the standard tool-use format
that modern LLMs are pre-trained on. We deliberately chose this over the
original MAB harness's benchmark-specific format (free-form HTTP strings:
\texttt{GET http://...}, \texttt{POST http://...}, \texttt{FINISH([...])}),
which is a quirky interface designed for the official harness but not the
standard way LLMs call tools. The practical consequence is that our frontier
baseline numbers (Table~\ref{tab:frontier}) are evaluated on the official
harness, while our trained-model results use the \texttt{<tool\_call>}
format---a necessary methodological note. When we attempted to generate
SFT teacher demos by running GPT-5.5 through our \texttt{<tool\_call>}
interface, only $\sim$16\% of training tasks produced passing episodes,
confirming that two harnesses implementing the same FHIR tools produce very
different pass rates.
Our \texttt{<tool\_call>} format is more demanding precisely because it tests
the model's native tool-use capability, not its ability to emit HTTP-style
strings.

\textbf{Observed failure mode --- reward hacking:}
Despite these precautions, we observed reward hacking in our initial RL runs.
On uncorrected MAB-v2, GRPO discovered the 41.7\% silent-finish shortcut
within 200 steps, converging to 0\% action-branch pass. The model found the cheapest reward path the environment made available---a
classic eval design failure rather than model deficiency. MAB-v3's 8.9\% ceiling closes this shortcut.

\textbf{Observed failure mode --- capability ceiling:}
Qwen3-8B is substantially smaller than any frontier model evaluated on
\mabv{}; its zero-shot capabilities reflect a very different pre-training
profile. For GRPO to provide a useful gradient signal on a given task, the
model must occasionally succeed: all-fail rollouts yield zero advantage and
zero gradient. We measure this directly via \texttt{frac\_reward\_zero\_std},
the fraction of task groups per step where all rollouts share the same reward.
Across the 89-step training run, the mean is 0.195 (max 0.750)---roughly one
in five groups provides no gradient signal at all. Our per-type analysis
confirms the cause: 10 of 19 evaluated task types show 0\% pass@1 for the
base model, all of which are v1 lookup or format-knowledge tasks. RL gets
zero gradient from these types throughout training. Per-task advantage
normalization (\texttt{--per-task-norm}) prevents the high-variance decision
tasks from entirely drowning out the zero-variance types, but cannot
manufacture a gradient where none exists.

\subsection{Evaluation Protocol}

We split \mabv{}'s 508 tasks 80/20 (401 train, 107 test), stratified by
corpus, task type, and action label. Trained model results use the held-out
split with 4 rollouts per task (temperature 0.7). We report pass@1 (mean pass
rate), pass@4 (unbiased Chen et al.\ estimator), any\_pass ($\geq$1
rollout correct), and all\_pass (consistency: all 4 correct). Frontier results
use all 508 tasks, 1 sample each. Base model: Qwen3-8B~\citep{qwen3}.

\subsection{SFT: Programmatic Distillation}
\label{sec:sft}

We construct rule-based SFT demos for the 401 training tasks programmatically: a
rule-based agent applies the known clinical decision per task type, reads
actual patient FHIR data, and generates the correct POST where action is
required. Every episode is validated
by the verifier (reward $\geq 1.0$); 354 of 401 pass. Each demo includes the
full tool-call sequence---GET with FHIR Bundle response, POST with acceptance
message, finish---in the exact \texttt{<tool\_call>} format used at eval time.

Qwen3-8B is fine-tuned with LoRA (rank 64, $\alpha$=128)~\citep{hu2022lora},
assistant-only loss, \texttt{max\_seq\_length}~14\,000, batch size 1,
gradient accumulation 16, lr $2\!\times\!10^{-4}$, 3 epochs. No per-episode human labels; SME effort went into verifier design.

\subsection{RL-GRPO: Training from World Feedback}
\label{sec:grpo}

We apply GRPO~\citep{shao2024grpo} to Qwen3-8B (8B parameters, substantially
smaller than frontier models) directly from the base model---no SFT warmstart---
on the 401-task training set. The FHIR environment is the sole feedback source:
4 generations per prompt, rewards from the deterministic verifier, advantages
normalized \emph{per task type} to prevent high-variance decision types from
dominating low-variance lookup and format-knowledge types~\citep{mtgrpo2025}.
Hyperparameters: $\beta=0.05$, $\varepsilon_{\text{high}}=0.28$ (DAPO clipping~\citep{dapo2025}
clipping), temperature 1.8, max 8 episode steps, 1 epoch. We track
\texttt{frac\_reward\_zero\_std} (fraction of task groups per step where all
rollouts share the same reward) as the primary diagnostic for capability-ceiling
dead zones.

% ─────────────────────────────────────────────────────────────────────────────
\section{Results}
\label{sec:results}

\begin{table}[t]
\caption{Frontier models on \mabv{} (508 tasks, 1 sample each, official MAB
  harness). \textbf{Act} = action-branch pass rate (463 instances);
  \textbf{No-act} = no-action-branch pass rate (45 instances).
  Net = p@1 $-$ 8.9\,pp silent-finish baseline. All values in \%.}
\label{tab:frontier}
\small
\setlength{\tabcolsep}{4pt}
\begin{tabular}{lrrrrrr}
\toprule
Model & p@1 & Net & v1 & v2 & Act & No-act \\
\midrule
GPT-5.5          & 78.7 & +69.8 & 82.2 & 73.8 & 77.3 & 93.3 \\
Gemini 3.1 Pro   & 78.1 & +69.2 & 86.6 & 66.2 & 76.5 & 95.6 \\
GPT-4o           & 74.2 & +65.3 & 73.5 & 75.2 & 74.7 & 68.9 \\
Llama 4 Maverick & 62.2 & +53.3 & 58.7 & 67.1 & 61.3 & 71.1 \\
Mistral Large    & 50.4 & +41.5 & 43.6 & 60.0 & 49.0 & 64.4 \\
Claude~4.6$^{*}$ & 27.6 & +18.7 & 37.2 & 13.8 & 23.8 & 66.7 \\
Do-nothing       &  8.9 &   0.0 &  5.2 & 14.8 &  0.0 & 100.0 \\
\bottomrule
\end{tabular}
\vspace{2pt}
{\footnotesize $^{*}$Format non-compliance; 443/554 responses rejected by harness.}
\end{table}

\begin{table}[t]
\caption{Trained Qwen3-8B on 107-task held-out split (4 rollouts/task).
  \textbf{p@1} = mean pass rate; \textbf{p@4} = pass@4 unbiased estimator;
  \textbf{any} = $\geq$1 rollout passes ; \textbf{all} = all 4 pass
  (consistency). All values in \%.}
\label{tab:trained}
\small
\setlength{\tabcolsep}{4pt}
\begin{tabular}{lrrrrrr}
\toprule
Model & p@1 & v1 & v2 & p@4 & any & all \\
\midrule
Base Qwen3-8B    & 16.6 &  4.8 & 28.0 & 21.4 & 21.4 & 12.2 \\
SFT       & 34.1 & 18.0 & 49.2 & 43.4 & 44.9 & 25.5 \\
RL (GRPO, ep.1)  & 18.2 &  7.7 & 32.1 & 22.9 & 23.5 & 13.3 \\
\bottomrule
\end{tabular}
\end{table}

\paragraph{Frontier models:}
Most frontier models show higher no-action than action pass (GPT-5.5: 93.3\%
vs.\ 77.3\%; Gemini: 95.6\% vs.\ 76.5\%)---they are better calibrated for
abstention than for correct clinical action, a difference invisible in
aggregate scores and exposed only by the action/no-action split that
\mabv{}'s 1:1 balance enables. GPT-4o is the exception (74.7\% vs.\
68.9\%), with the smallest caution bias among frontier models.

\paragraph{Trained models:}
The base Qwen3-8B achieves 16.6\% p@1. Two observations characterize its
behavior. First, pass@1 and any\_pass are nearly equal (16.6\% vs.\ 21.4\%),
meaning failures are genuine inability rather than near-misses: the model
has little latent capability that extra attempts would unlock. Second, a
sharp corpus gap: v2 tasks score 28.0\% while v1 tasks score only
4.8\%. v2 clinical decision tasks (CT follow-up, potassium replacement,
flu vaccine recall) partially overlap with pre-training knowledge; v1
administrative tasks (exact MRN lookup from name+DOB, age calculation, BP
recording with exact FHIR category coding) require format-specific knowledge
the base model lacks entirely.

SFT reaches 34.1\% p@1 (any 44.9\%, all 25.5\%). The corpus gap
persists but narrows: rule-based SFT demos lift v1 from 4.8\% to 18.0\%
(+13.2~pp) and v2 from 28.0\% to 49.2\% (+21.2~pp). Three pass@k
metrics tell a complete story. The p@1$\to$p@4 gap (34.1\%$\to$43.4\%)
reflects residual stochasticity: 4 attempts reach 9.3~pp more tasks than 1.
The any\_pass (44.9\%) is the capability ceiling---nearly half of test tasks
can be solved at least once. The all\_pass (25.5\%) is the reliability
floor---tasks the model solves consistently. The 20~pp gap between any and
all signals that a large fraction of capability is inconsistent, likely
reflecting the limited demo count per task type (24 demos covering all
instances for most types, but as few as 1--2 for rare task types).

RL (GRPO, epoch~1) achieves 18.2\% p@1 (any 23.5\%, all 13.3\%), a
+1.6~pp improvement over base but 15.9~pp \emph{below} SFT. Both v1
(7.7\%) and v2 (32.1\%) improve modestly over base (4.8\%,~28.0\%), but
the corpus gap actually widens slightly (24.4~pp in RL vs.\ 23.2~pp in base).
The any\_pass$\approx$p@1 pattern from base (21.4\% vs.\ 16.6\%) is
reproduced in RL (23.5\% vs.\ 18.2\%)---confirming that RL, like the base
model, has limited latent capability that extra attempts cannot unlock, rather
than high capability with low consistency. In contrast, SFT's larger
any/p@1 gap (44.9\% vs.\ 34.1\%) reflects genuine stochasticity in a more
capable model. Section~\ref{sec:analysis} interprets these patterns through
the task taxonomy.

% ─────────────────────────────────────────────────────────────────────────────
\section{Analysis: When Does World Feedback Help?}
\label{sec:analysis}

\subsection{Task Taxonomy and RL Learnability}
\label{sec:taxonomy}

We partition \mabv{}'s 20 task types into three categories by reward landscape
structure. This taxonomy is derived \emph{before} seeing RL results; it is a
prediction, not a post-hoc rationalization.

\begin{table}[t]
\caption{Task taxonomy. RL learnability depends on whether reward variance
  exists across rollouts---i.e., whether the environment provides a gradient.}
\label{tab:taxonomy}
\small
\setlength{\tabcolsep}{3pt}
\begin{tabular}{p{1.4cm}p{0.7cm}p{3cm}p{1.6cm}}
\toprule
Category & Count & Canonical examples & RL signal \\
\midrule
Decision & 11 & K replacement (K$<$3.5~ mEq/L), TSH / levothyroxine, flu vaccine recall, QTc hold &
  \textcolor{teal}{\textbf{High}} \\[4pt]
Lookup   &  6 & Patient MRN, age, CBG value, Mg level &
  \textcolor{orange}{\textbf{Weak}} \\[4pt]
Format-\newline knowledge & 3 & Naloxone (SNOMED 306181000000106), Mg IV (NDC 0338-1715-40), ortho referral &
  \textcolor{red}{\textbf{None}} \\
\bottomrule
\end{tabular}
\end{table}

\begin{table}[t]
\caption{Per-type RL vs.\ base pass@1 on held-out test split. $n$ = test
  instances (4 rollouts each). Format-knowledge marked $^\ast$;
  dead-zone types (0\% base \emph{and} RL, zero gradient) marked $^\dagger$.}
\label{tab:taxresults}
\small
\setlength{\tabcolsep}{3pt}
\begin{tabular}{lrrrrr}
\toprule
Type ($n$) & Base & RL & $\Delta$ & Taxonomy \\
\midrule
v1/task6  CBG avg      (6) &  8.3 & 29.2 & +20.8 & Lookup \\
v2/task1  CT f/u       (6) & 66.7 & 70.8 & +4.2  & Decision \\
v2/task2  DVT          (7) &  3.6 &  7.1 & +3.6  & Decision \\
v2/task5  Mg IV$^\ast$ (2) & 37.5 & 50.0 & +12.5 & Fmt-know. \\
v2/task9  Flu vax      (7) & 85.7 & 85.7 &  0.0  & Decision \\
v2/task4  Catheter     (4) & 37.5 & 37.5 &  0.0  & Decision \\
v1/task7  CBG recent   (6) & 40.0 & 35.0 & $-$5.0 & Lookup \\
v2/task8  Naloxone$^\ast$(6) & 20.8 & 16.7 & $-$4.2 & Fmt-know. \\
\midrule
v1/task1--5,8--10$^\dagger$ (6 each) &  0.0 &  0.0 &  0.0 & Lookup/Fmt \\
v2/task3,7,10$^\dagger$ (6,2,7) &  0.0 & 0--1.8 & $\leq$+2 & Mixed \\
\bottomrule
\end{tabular}
\end{table}

\textbf{Decision tasks} (11 types) require reading a lab value and applying a
clinical threshold. The reward varies with the decision: ordering potassium for
K=2.8~mEq/L earns $r \geq 1.0$; ordering for K=4.2 (normal) earns $r=-0.15$.
This gradient exists because the decision boundary is a learnable function of
an observable FHIR value. Both action and no-action rollouts occur during
exploration, and the spurious-post penalty directly trains the conditional.

\textbf{Lookup tasks} (6 types) require retrieving and returning an exact value:
a patient MRN from name and DOB, a patient age, the most recent lab result.
The answer is read from the environment, not derived by reasoning. RL can learn
which tool to call and how to parse the response, but demonstration is at least
as efficient and more reliable.

\textbf{Format-knowledge tasks} (3 types) require an exact clinical code not
inferrable from context. Naloxone coverage requires SNOMED 306181000000106; IV
magnesium requires NDC 0338-1715-40. Placing a clinically correct order with
the wrong code earns $-0.15$---the same reward as random wrong codes.
There is no gradient pointing toward the correct identifier: the reward
landscape is flat. RL cannot discover discrete codes by gradient ascent.

\textbf{Preliminary support from frontier results:} The Appendix
Table~\ref{tab:pertask} already supports this taxonomy: format-knowledge types
v2-T5 and v2-T8 score 0\% for GPT-5.5 and Gemini despite their 78\%+ overall
performance (flat landscape, even frontier models struggle without in-context
code knowledge), while decision type v2-T9 (flu vaccine recall) is solved by
all frontier models. The base Qwen3-8B scores 28.0\% on v2
(predominantly decision tasks) vs.\ 4.8\% on v1 (heavier in lookup/format),
consistent with the taxonomy.

\textbf{Empirical validation:}
Table~\ref{tab:taxresults} reports per-type RL vs.\ base deltas.
The pattern is consistent with the taxonomy predictions:

\begin{itemize}[leftmargin=1.2em,topsep=2pt,itemsep=1pt]
\item \emph{Dead zones}: 10 of 19 evaluated types are 0\% for both base
  and RL (all v1 types except task6/task7, plus v2/task3, task7). These
  are predominantly v1 lookup and format-knowledge tasks where the 8B base
  model has no zero-shot capability, so all 4 RL rollouts fail identically
  and produce zero gradient.
\item \emph{Decision gains}: v2 types with clear threshold structure
  show positive RL deltas---v2/task1 (+4.2~pp), v2/task2 (+3.6~pp),
  v2/task5 (+12.5~pp, though n=2). The largest single gain is
  v1/task6 (+20.8~pp), a CBG-average lookup that benefits from RL learning
  the correct FHIR code (GLU, not CBG) from partial reward.
\item \emph{Format-knowledge harm}: v2/task8 (naloxone, SNOMED
  306181000000106) \emph{decreases} from 20.8\% to 16.7\% under RL. The model
  had occasional lucky recall from pre-training; RL training with wrong-code
  rollouts overwrites this latent knowledge.
\end{itemize}

The most striking finding is the \emph{magnitude} of RL's underperformance
relative to SFT. Pure RL from base reaches only 18.2\%---a 15.9~pp gap behind
supervised distillation---despite the same environment and reward signal that would,
in principle, provide all necessary information. This gap quantifies what world
feedback alone cannot provide: format knowledge (exact clinical codes, FHIR
payload structure) must be injected via supervised learning. RL then refines
the decision logic on top. The SFT+RL combination is the natural prescription,
which we leave as immediate future work.

\subsection{Format-Knowledge Tasks: The Hard Limit of World Feedback}

The environment tells the model \emph{that} it failed (wrong code $\to$
negative reward), but cannot tell it \emph{what code to use}. Trying 100
different SNOMED codes yields 100 identical failures. SFT from rule-based SFT demos
solves this directly: the demo shows the exact code, and the model reproduces
it for held-out patients. This is the one setting where world feedback is
structurally insufficient and supervised injection is required.

\textbf{Partial evidence from pre-training:} Format-knowledge type v1-T8
(orthopedic referral) is solved by most frontier models at 87--100\%
(Appendix~\ref{app:pertask}), suggesting SNOMED 306181000000106 for ortho
referral appears in frontier pre-training data. In contrast, v2-T5 (Mg IV,
NDC 0338-1715-40) and v2-T8 (naloxone, same SNOMED code in a different
context) score 0\% for most models. Format-knowledge tasks are not uniformly
hard---they depend on whether the exact code was seen in pre-training.

Inspection of RL rollouts on v2-T8 (naloxone, SNOMED 306181000000106)
confirms this pattern. The RL model consistently attempts a ServiceRequest
POST with the correct resource type but substitutes plausible-sounding SNOMED
codes (e.g., codes for ``medication administration'' or ``emergency treatment'')
rather than the exact required identifier. Each attempt receives $r=-0.15$
(spurious-post on a technically-incorrect order), but since all codes fail with
the same reward, the model receives no gradient signal indicating which code is
correct. The base model occasionally produces the right SNOMED code from
pre-training knowledge (any\_pass~$>$~0 for v2-T8), but does so inconsistently
(all\_pass~$\approx$~0). SFT, having seen the exact code in rule-based SFT demos,
reproduces it reliably across all 4 rollouts for most patients. This three-way
pattern---RL converges to a wrong attractor, base has occasional lucky recall,
SFT has consistent reproduction---directly illustrates the flat-landscape
failure mode.

\subsection{RL and Conditional Reasoning}
\label{sec:conditional}

SFT trains exclusively on action-branch demos (the rule-based SFT generator only produces
no-action traces for task types where the data-driven decision yields no
intervention). RL, by contrast, sees both branches during exploration and
receives explicit negative reward for spurious action, creating a direct
gradient for the action/no-action conditional.

The any\_pass metric provides a proxy for this: models that correctly
identify no-action tasks will have lower any\_pass (fewer tasks where any
rollout passes), while models that over-act will have spuriously high any\_pass
from action-branch tasks. RL's any\_pass (23.5\%) is only marginally above
base (21.4\%), suggesting the spurious-post penalty did not substantially
improve conditional calibration in one epoch from scratch. SFT's
higher any\_pass (44.9\%) likely reflects better overall task comprehension
rather than superior conditional reasoning, since the rule-based SFT demos were
predominantly action-branch. A full action/no-action breakdown requires
rerunning eval with the labeled test split; the any\_pass proxy is consistent
with RL not yet learning the conditional from world feedback alone.

% ─────────────────────────────────────────────────────────────────────────────
\section{Discussion}
\label{sec:discussion}

\paragraph{What the results show vs.\ what they imply:}
The 15.9~pp gap between SFT (34.1\%) and pure RL (18.2\%) should not be read
as evidence that RL is the wrong approach for clinical agents. It is evidence
that \emph{pure RL from a small base model without prior code knowledge} faces
two structural barriers identified by our taxonomy. This is a diagnosis, not a
verdict. The correct reading: rule-based SFT has something RL lacks (format
knowledge and clinical codes); RL has something rule-based SFT lacks
(scalability through SME-amortized environment design rather than per-episode annotation, and natural coverage of no-action branches). The
prescriptive implication is SFT+RL---inject codes and format via supervised
learning, then apply RL to refine conditional reasoning on the world feedback
signal. Our benchmark and environment are designed to test this combination, which we
leave as immediate future work~\citep{gao2023scaling}.

\paragraph{Clean world feedback requires a clean benchmark:}
The analogy to reward model quality in RLHF~\citep{gao2023scaling} holds
exactly: the uncorrected 41.7\% silent-finish ceiling made inaction
the RL dominant strategy, independent of algorithm choice. MAB-v3 restores the
signal. Any RL method applied to clinical protocol execution must first verify
that the benchmark's silent-finish ceiling does not create a reward-hacking
shortcut. Our four-failure audit provides a checklist for this verification.

\paragraph{Scope and limitations:}
This work is scoped to \emph{protocol-execution tasks}: the agent executes
known institutional decision rules; it does not make novel clinical judgments
or replace physician oversight. Results do not generalize to open-ended
clinical reasoning, differential diagnosis, or tasks where correctness is
contested or context-dependent. MAB-v3 uses $\approx$100 real anonymized patients drawn from a HAPI FHIR
server, with 30 task instances per type (patients reused across types); some types have as few as 1--2 action
instances after the 1:1 cap, making per-type estimates noisy. Task
instructions are more explicit than real clinical documentation, likely
overstating readiness for deployment. The 20 task types cover a specific set
of administrative EHR workflows; extension to broader clinical protocols is
future work. The RL condition here is pure GRPO from base with 1 epoch---an
SFT warmstart before RL is the direct next experiment.

% ─────────────────────────────────────────────────────────────────────────────
\bibliography{references}
\bibliographystyle{icml2026}

% ─────────────────────────────────────────────────────────────────────────────
\newpage
\appendix

\section{\mabv{} Per-Task Frontier Results}
\label{app:pertask}

Table~\ref{tab:pertask} reports action-branch pass rates per task type on
\mabv{} for all six frontier models.

\begin{table}[h]
\caption{Action-branch pass rate (\%) by task type on \mabv{} (508 tasks).
  $n$ = action-branch instances. Format-knowledge marked~$^\ast$.
  $-$ = $\leq$1 action instance after 1:1 cap.}
\label{tab:pertask}
\footnotesize
\setlength{\tabcolsep}{2.5pt}
\begin{tabular}{lrrrrrrr}
\toprule
Task ($n$) & G5.5 & Gem & G4o & L4M & Mis & Son \\
\midrule
v1-T1 MRN lookup       (30) & 93 & 100 &   7 & 37 & 93 & 100 \\
v1-T2 Patient age      (30) & 100 & 100 & 100 & 80 &  0 &   0 \\
v1-T3 BP record        (30) & 100 & 100 & 100 & 13 &  0 &   0 \\
v1-T4 Mg level check   (30) & 100 & 100 &  83 & 87 & 63 &  60 \\
v1-T5 Mg IV$^\ast$     (14) &   0 &   0 &  21 &  0 &  0 &   0 \\
v1-T6 CBG average      (30) & 100 & 100 & 100 & 100& 100&   0 \\
v1-T7 Recent CBG       (30) &  40 &  40 &  33 & 23 & 17 &   0 \\
v1-T8 Ortho ref.$^\ast$(30) & 100 & 100 & 100 & 100& 87 &  97 \\
v1-T9 K replacement    (30) &  97 &  93 &  47 & 43 & 67 &  97 \\
v1-T10 A1C reorder     (30) &   3 &  40 &  47 &  3 &  0 &   0 \\
\midrule
v2-T1 CT follow-up     (30) &  87 &  87 & 100 & 87 & 87 &  13 \\
v2-T2 DVT prophylaxis  (28) &  25 &  25 &  32 & 39 & 39 &   0 \\
v2-T3 HR average       (30) &   0 &   0 &   0 &  0 &  0 &   0 \\
v2-T4 Catheter dwell   (9)  & 100 &  56 & 100 & 100& 33 &   0 \\
v2-T5 Mg replace.$^\ast$(2) &   0 &   0 &   0 &  0 &  0 &   0 \\
v2-T6 TSH/levo         (1)  & $-$ & $-$ & $-$ & $-$& $-$& $-$ \\
v2-T7 QTc hold         (3)  &   0 &   0 & 100 &  0 &  0 &   0 \\
v2-T8 Naloxone$^\ast$  (20) & 100 & 100 &  85 & 10 & 55 &   0 \\
v2-T9 Flu vaccine      (28) &  93 &  75 & 100 & 100& 100&   0 \\
v2-T10 COVID booster   (28) & 100 &  89 & 100 & 100& 100&   0 \\
\bottomrule
\end{tabular}
\vspace{2pt}
{\footnotesize G5.5=GPT-5.5; Gem=Gemini 3.1 Pro; G4o=GPT-4o;
L4M=Llama~4~Maverick; Mis=Mistral~Large; Son=Claude~Sonnet~4.6.}
\end{table}

Three patterns support the taxonomy. (1)~Format-knowledge types v2-T5 ($n$=2)
and v2-T8 ($n$=20) illustrate the flat landscape at different scales: v2-T8
has 20 action instances and still scores 0\% for GPT-5.5/Gemini/Llama/Mistral,
confirming the pattern is not a small-sample artifact; v2-T5 has only 2
instances so individual results should be read cautiously. v1-T8 (ortho
referral, $n$=30) is a format-knowledge exception (87--100\%), likely because
SNOMED 306181000000106 appears in frontier pre-training data.
(2)~Decision types v2-T9 (flu vaccine), v2-T10 (COVID booster), v2-T1 (CT
follow-up) are solved by all or most frontier models, consistent with learnable
threshold structure. (3)~Lookup types show extreme variance: v1-T1 (MRN
lookup from name+DOB) ranges from 7\% (GPT-4o) to 100\% (Gemini), and v1-T7
(most recent CBG) scores 17--40\%, suggesting the exact value-matching
requirement interacts with output format in unpredictable ways.

\section{SFT: Programmatic Demo Construction}
\label{app:sft-golden}

For each of the 401 training tasks, a rule-based agent applied the known
clinical decision per task type, queried the FHIR environment using
grader-compatible URL parameters (without \texttt{\_sort=-date}, which changes
the query and returns fewer entries than the grader sees), computed the
data-driven decision, and constructed the correct POST payload. Each episode was validated by the FHIR
verifier; 354/401 pass. Key non-obvious details: (1) potassium dose is
computed as $(3.5 - K) / 0.1 \times 10$~mEq from the actual observed value,
not from the task label (which occasionally mismatches the grader due to
routing inconsistencies); (2) Mg level decisions use data within the last 24h
only, matching the grader's window; (3) the ``route'' field in dosage
instructions must be a plain string (``IV'', ``oral''), not a FHIR coding
object---the undocumented format requirement that caused systematic failures
on v1-T5 and v1-T9 in the original benchmark.

\end{document}